\documentclass[runningheads]{llncs}

\usepackage[T1]{fontenc}
\usepackage{graphicx}
\usepackage{xcolor}
\usepackage{amsmath}
\usepackage{amssymb}
\usepackage{algorithm}
\usepackage{algpseudocode}
\usepackage{booktabs}
\usepackage{multirow}
\usepackage[misc]{ifsym}
\usepackage{bbding}
\usepackage{amsfonts}
\usepackage[bottom]{footmisc}
\usepackage[hyphens]{url}
\usepackage[hidelinks,breaklinks]{hyperref}
\usepackage{orcidlink}

\begin{document}

\title{Self-Supervised Bio-Inspired Robotic Trajectory Planning with Obstacle Avoidance}
\titlerunning{Self-Supervised Robotic Trajectory Planning with Obstacle Avoidance}
\author{
    Miroslav Krupa\inst{1} \and
    Miroslav Cibula\inst{1}\orcidlink{0009-0000-8460-6533} \and
    Krist\'ina Malinovsk\'a\inst{1}\textsuperscript{(\Letter)}\orcidlink{0000-0001-7638-028X}
}
\authorrunning{M.~Krupa et al.}
\institute{
    Faculty of Mathematics, Physics and Informatics, Comenius University Bratislava, Bratislava, Slovakia \\
    \email{krupa39@uniba.sk, cibula25@uniba.sk, kristina.malinovska@fmph.uniba.sk}
}

\maketitle

\begin{abstract}
Trajectory planning is a fundamental problem in robotics, requiring the generation of collision-free and efficient trajectories in a potentially complex environment. While sampling-based planners remain the dominant approach, they are often computationally expensive, particularly in high-dimensional spaces and obstacle-rich environments. Methods based on model learning offer a promising alternative, enabling efficient planning through a bounded number of forward passes through a neural trajectory planner, but commonly suffer from low sample efficiency or limited generalisation due to their reliance on exploration or expert demonstrations. 
This follow-up work tests our neuro-inspired self-supervised learning framework for trajectory planning that leverages forward and inverse models as the internal supervisory mechanism in an environment that contains an obstacle.
Experimental results demonstrate the feasibility of the approach while revealing a tendency of our planner to exploit the learning signal provided by the forward and inverse models. To address this issue, additional training regimes and mitigation strategies are proposed and evaluated.
\end{abstract}

\section{Introduction}

Motion planning, i.e., finding the optimal path between two poses of a robot or its manipulator, remains one of the fundamental and relevant challenges in robotics \cite{Dobis2022}. A trajectory planner is an algorithm that computes a collision-free sequence of states and actions to safely transition from the robot's initial position to the specified goal position, particularly in obstacle-rich environments. Sampling-based planning methods represent the current state of the art in robotic motion planning \cite{Karaman2011}. These methods iteratively construct a graph by connecting a sample produced by a sampling strategy from the specified planning space (e.g., a Cartesian space of end-effector positions or a space of the robot's joint-angle configurations) to the existing graph via a local planner. A feasible trajectory is then obtained by finding a path from the start to the goal node. However, the methodology frequently requires post-processing, such as smoothing, to produce natural motion, and its performance deteriorates in high-dimensional planning spaces and obstacle-rich environments. For complex humanoids, this can lead to unpredictable run times and difficulty finding a near-optimal solution.

Alternative methods for trajectory planning include learning-based approaches \cite{Hu2026}, which learn motion strategies from experience or demonstration, such as reinforcement learning and imitation learning. While the former allows discovering viable behaviour through interaction with the environment, it often suffers from sample inefficiency and typically requires extensive exploration, which may be hazardous or costly, posing challenges when deploying in physical robotic systems. In contrast, imitation learning avoids the need for exploration by leveraging expert demonstrations, but its performance is strictly tied to the quality of the training dataset, which may limit generalisation and prevent the discovery of optimal behaviour. To address these limitations we proposed a novel neuro-inspired self-supervised learning framework that incorporates separately trained forward (FM) and inverse models (IM) into the training of a trajectory planner \cite{Cibula2025}. The FM and IM, analogous to cognitive internal sensorimotor models, learn the possible motions constrained by the robot's kinematics. This internalised knowledge is then utilised to train a trajectory planner that can produce trajectories within a fixed runtime, making it highly efficient when many trajectories need to be planned in the same environment. 

In the present article, we extend our neural trajectory planner \cite{Cibula2025} to respond to obstacles in the environment and plan the trajectories while avoiding the collision with them. To this end, the FM and IM are retrained to account for the presence of an obstacle in the environment. The models are subsequently used in the trajectory planner's training to correct its predicted trajectories. The empirical error between the predicted and corrected trajectory serves as self-supervised learning feedback, guiding the trajectory model (TM) to generate feasible trajectories. We present an evaluation of an industrial 7-degree-of-freedom robotic arm in a simulated environment with a single obstacle, and interpret our results as another step towards constant-runtime neural trajectory planners.

\section{Related Work}
\label{sec:related}

Sampling-based planning begins by constructing a graph or tree from samples of the robot's collision-free configuration space, with nodes connected via local steering to approximate paths in the high-dimensional spaces. Many classical algorithms \cite{Karaman2011}, such as probabilistic roadmaps and rapidly-exploring random trees, are probabilistically complete if a feasible path exists.  However, these approaches are not asymptotically optimal, meaning there is no guarantee of path quality, which contradicts the aim of planning smooth, efficient trajectories, and while probabilistic completeness ensures convergence in probability, the process can be computationally intensive and overly demanding on memory.
A recent comparative review \cite{Karaman2011} shows that modern implementations of these sampling-based algorithms balance success rate, runtime, and path quality, and that performance varies across scenarios such as passage widths and the number of robot degrees of freedom. Although this category of planners contains some disadvantages, it is evident that they remain the workhorse of complex motion planning.

In reinforcement learning, the trajectory is viewed as a sequence of the environment's states and actions performed by the robot under a policy learned through interaction with the environment, with the aim of maximising cumulative reward by completing the motion task \cite{Xie2019}. Such models have shown promise for obstacle-avoiding path planning and tracking that respect the robot's environmental constraints \cite{ElgueaAguinaco2024}.
However, because reinforcement learning methods generally entail substantial data requirements and exploration costs, offline forms of this paradigm \cite{FigueiredoPrudencio2024}, as well as combinations with supervised approaches, have been explored to leverage the advantages of demonstration learning \cite{Emmons2022}. 

Issues such as reward design and sample efficiency have motivated alternative formulations of reinforcement learning problems, including supervised sequence modelling approaches, in which task solving is reformulated as generating temporally ordered states, actions, and rewards, constituting trajectories \cite{Janner2021}.
Such approaches mitigate the above-mentioned disadvantages of reinforcement learning while capitalising on the strengths of sequence modelling, such as improved scalability and adaptable representations provided by recurrent neural networks or transformer-based architectures. Nevertheless, these implementations tend to suffer from common supervised-learning limitations, a higher failure rate outside the training distribution, and low explainability and verifiability due to their end-to-end system design \cite{Hu2026}.
Our previous work \cite{Cibula2025} presents a novel bio-inspired self-supervised approach to trajectory planning, utilizing paired forward and inverse models and a recurrent predictor. Rather than relying on an explicit training distribution, the FM and IM enrich the training process by providing more informative learning targets. Our model overcomes some of the abovementioned problems and generally aims at improving the generalisation ability and the feasibility of the generated trajectories.

\section{Methods}
\label{sec:methods}

This work aims to implement, train, and evaluate a neural trajectory planner that receives vector representations of the initial and the goal state of the environment and generates a sequence of intermediate states facilitating transitions between them. The generated trajectory enables the robot to perform a motion task by reaching the desired goal state from the initial state. Depending on the features of the representation used, the trajectory may consist of either joint configurations directly followed by the robot or end-effector positions, from which joint configurations are computed through an inverse kinematics solver.
The planner follows a self-supervised learning paradigm in which a trajectory model is trained with the aid of separately trained forward and inverse models. In this section, we describe the architectures of the individual components and their corresponding training procedures. The overall design and training approach are adapted from our trajectory model \cite{Cibula2025}.

To assess the robustness of the architecture, two datasets are generated using context-specific representations of the state of the environment at time $t$, vector $\mathbf{s}_t$, and of the action taken or control input followed by a robot at time $t$, vector $\mathbf{a}_t$. 
First, we generate a transition dataset for imitation learning of the FM and the IM, where each data point is represented by a triplet $\left(\mathbf{s}_t, \mathbf{a}_t, \mathbf{s}_{t + 1}\right)$, denoting an event of performing $\mathbf{a}_t$ in $\mathbf{s}_t$, resulting in $\mathbf{s}_{t + 1}$. Secondly, a trajectory dataset is generated for training the TM. Each trajectory is defined as $\tau = \left(\mathbf{s}_0, \mathbf{a}_0, \mathbf{s}_1, \mathbf{a}_1, \mathbf{s}_2, \ldots, \mathbf{s}_{T - 1}, \mathbf{a}_{T - 1}, \mathbf{s}_T\right)$, where $\mathbf{s}_0$ denotes the initial state, $\mathbf{s}_T$ the final (goal) state, and $T$ the trajectory length.
The data generation procedures used to obtain these datasets implement the approach proposed in \cite[Sec.~4.1]{Krupa2026}.

The FM is learned by a multi-layer perceptron $\rm FM$ that approximates the system dynamics denoted by $\it fm$, such that
\begin{equation}
    \text{FM}\!\left(\mathbf{s}_t, \mathbf{a}_t \right) \equiv \mathbf{\hat{s}}_{t + 1} \approx \mathbf{s}_{t + 1} \equiv \mathit{fm} \!\left(\mathbf{s}_t, \mathbf{a}_t \right),
\end{equation}
where $\mathbf{\hat{s}}_{t + 1}$ is the predicted next state, which is an approximation of the ground-truth next state $\mathbf{s}_{t + 1}$. The output is structured into multiple heads, one for each subvector of the state vector. This design allows individual state components, such as the robot's joint-angle configuration and the obstacle's position and rotation, which differ in scale and semantic meaning, to be optimised more effectively.
During the FM training, the loss function
\begin{equation}
    \mathcal{L}_{\mathrm{FM}} \equiv \mathcal{L}_{\mathbf{s}} \!\left(\mathbf{\hat{s}}_{t + 1}, \mathbf{s}_{t + 1} \right) 
        \triangleq \frac{1}{k} \sum_{\mathbf{y}^{(i)}_{t + 1} \subseteq \mathbf{s}_{t + 1},\ \mathbf{\hat{y}}^{(i)}_{t + 1} \subseteq \mathbf{\hat{s}}_{t + 1}} 
            \mathcal{L}_{\mathbf{y}^{(i)}} \!\left(\mathbf{\hat{y}}^{(i)}_{t + 1}, \mathbf{y}^{(i)}_{t + 1} \right)
    \label{eq:Ls}
\end{equation}
is minimised. The constant $k$ represents the number of output heads, $\mathbf{y}^{(i)}_{t + 1}$, $\mathbf{y}^{(i)}_{t + 1} \subseteq \mathbf{s}_{t + 1}$ the $i$-th subvector of the vector $\mathbf{s}_{t + 1}$, and $\mathcal{L}_{\mathbf{y}^{(i)}}$ the loss function used for the subvector $\mathbf{y}^{(i)}_{t + 1}$. The loss is then formulated as a weighted sum of the prediction errors of each head of the network.

Analogously, the IM is learned by a multi-layer perceptron $\rm IM$ that approximates the inverse dynamics $\it im$, such that
\begin{equation}
\mathrm{IM} \!\left(\mathbf{s}_t, \mathbf{s}'_{t + 1} \right) \equiv \mathbf{\hat{a}}_t \approx \mathbf{a}_t \equiv \mathit{im} \!\left(\mathbf{s}_t, \mathbf{s}'_{t + 1} \right),
\end{equation}
where $\mathbf{\hat{a}}_t$ is the predicted action causing the transition between state $\mathbf{s}_t$ and the reduced representation $\mathbf{s}'_{t + 1}$ of the next state $\mathbf{s}_{t + 1}$, where components directly resulting from the action taken are excluded.
During the IM training, the following loss function is minimised
\begin{equation}
    \mathcal{L}_{\mathrm{IM}} \triangleq \mathcal{L}_{\mathbf{a}}(\mathbf{\hat{a}}_t, \mathbf{a}_t)
\end{equation}

\begin{figure}[t]
    \centering
    \includegraphics[width=\textwidth]{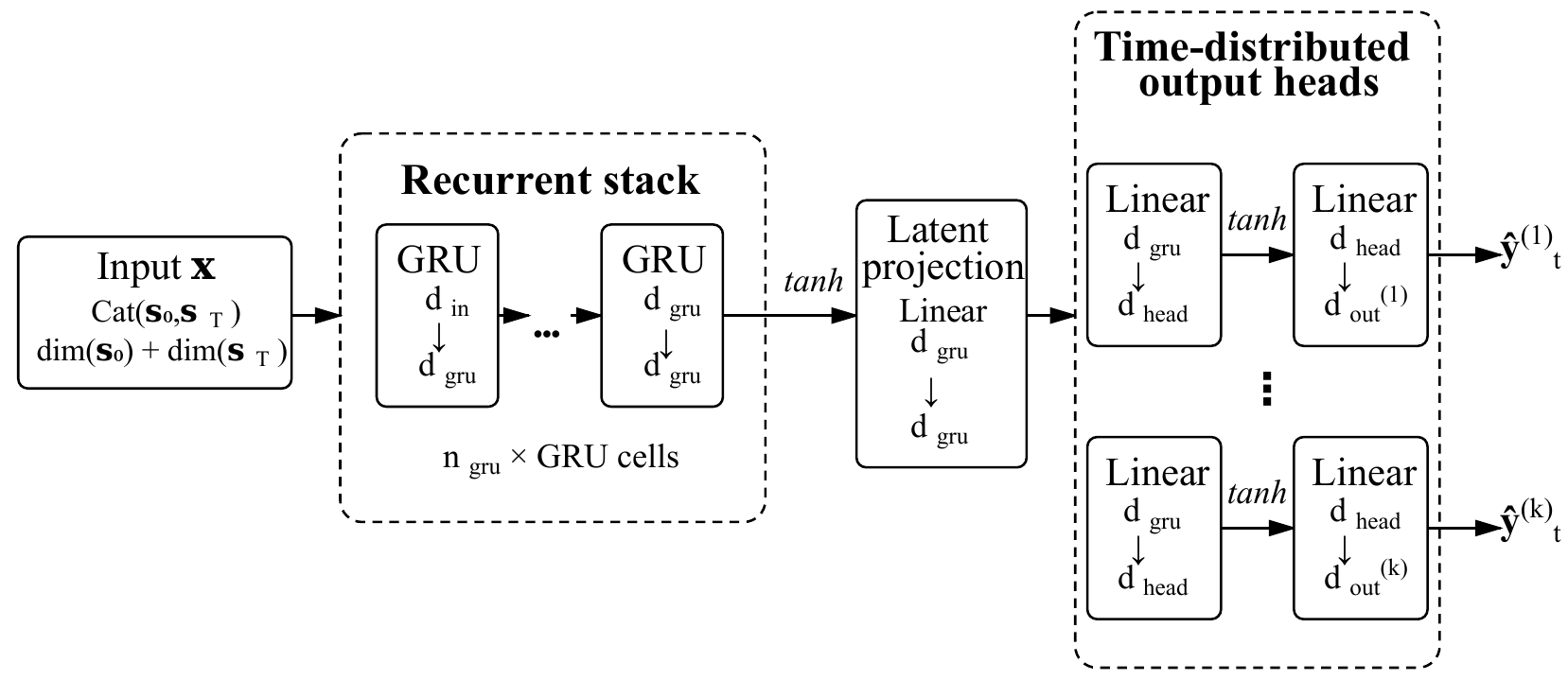}
    \caption[Architecture of the trajectory model]
    {Architecture of the trajectory model. At each time step
    , the same GRU stack is recursively executed, and the time-distributed output heads produce 
    the trajectory $\hat{\tau}_{\mathbf{s}}$.
    Dimensions of layers are denoted as $d_x$, where $x$ is the respective model part. $n_{\mathrm{gru}}$ is the number of GRUs.}
    \label{fig:tm_architecture}
\end{figure}

Finally, the trajectory model $\rm TM$ is a neural network with recurrent layers implemented as gated recurrent units (GRU) \cite{Cho2014}, defined as
\begin{equation}
    \mathrm{TM} \!\left(\mathbf{s}_0, \mathbf{s}_T \right) \equiv \hat{\tau}_\mathbf{s} \equiv \left( \mathbf{\hat{s}}_1, \ldots, \mathbf{\hat{s}}_{T - 1} \right),
\end{equation}
where $\hat{\tau}_\mathbf{s}$ is the predicted trajectory represented as a sequence of state vectors. During inference, the model receives an input consisting of the initial state $\mathbf{s}_0$ and the desired goal state $\mathbf{s}_T$, which the generated trajectory $\hat{\tau}_\mathbf{s}$ connects. Similarly to the FM, the time-distributed output of the recurrent network is divided into multiple output heads, where each head corresponds to a separate component of the state vector (see Fig.~\ref{fig:tm_architecture}).
The model's output $\hat{\tau}_\mathbf{s}$ is used together with the initial state $\mathbf{s}_0$ and the final state $\mathbf{s}_T$ of the ground-truth trajectory for the rectification process. Given the predicted trajectory, the algorithm produces a rectified trajectory
\begin{equation}
    \tilde{\tau}_{\mathbf{s}} = \left(\tilde{\mathbf{s}}_0, \tilde{\mathbf{s}}_1, \dots, \tilde{\mathbf{s}}_{T-1}, \tilde{\mathbf{s}}_T\right),
\end{equation}
consisting of rectified state vectors. The first and final states are fixed to the ground-truth states, meaning $\tilde{\mathbf{s}}_0 = \mathbf{s}_0$ and $\tilde{\mathbf{s}}_T = \mathbf{s}_T$. For each intermediate time step $t$, the IM predicts an action $\hat{a}_{t-1}$ from the previous rectified state $\tilde{\mathbf{s}}_{t-1}$ and the reduced predicted state $\hat{\mathbf{s}}'_t$, obtained from $\hat{\mathbf{s}}_t$ by removing components directly determined by the action vector. The FM then predicts the next rectified state $\tilde{\mathbf{s}}_t$ using the previous rectified state and the predicted action.

The rectified trajectory $\tilde{\tau}_{\mathbf{s}}$ is used in the training of the TM, allowing the FM and IM to provide a supervisory feedback signal. The result of such training is presumed to be a planner generating internally consistent and feasible trajectories that respect the system's dynamics and environmental constraints (under the assumption that the models responsible for trajectory corrections approximate the dynamics sufficiently well). The TM is trained by minimising the loss function
\begin{equation}
    \mathcal{L}_{\mathrm{TM}}(\hat{\tau}_s, \tilde{\tau}_s, \mathbf{s}_0, \mathbf{s}_T) 
        \triangleq \frac{1}{|\hat{\tau}_\mathbf{s}|} \sum_{t = 1}^{T - 1} 
            \mathcal{L}_\mathbf{s} \!\left(\mathbf{\hat{s}}_t, \mathbf{\tilde{s}}_t \right) + \mathcal{L}_{\rm init} \!\left(\mathbf{\hat{s}}_1, \mathbf{s}_0 \right) + \mathcal{L}_{\rm goal} \!\left(\mathbf{\hat{s}}_{T - 1}, \mathbf{s}_T \right),
    \label{eq:LTM}
\end{equation}
where the first term represents the weighted rectification error, encouraging the predicted trajectory to remain feasible and consistent regarding the learned FM and IM. 
The remaining terms guide the model towards intermediate states that form a valid connection between the initial and goal states.

\section{Experiments and Results}\label{experiments}

In this work, we define the state vector
\begin{equation}
    \mathbf{s}_t \triangleq \left(\boldsymbol{\theta}_t, {\it mgt}_t, \mathbf{ef}_{\mathit{xyz}, t}, \mathbf{ef}_{R, t}, \mathbf{g}_{\mathit{xyz}, t}, \mathbf{g}_{R, t}, \mathbf{o}_{\mathit{xyz}, t}, \mathbf{o}_{R, t} \right),    
    \label{eq:state_vector}
\end{equation}
where $\boldsymbol{\theta}_t \in \mathbb{R}^7$ is the joint-angle configuration of the robot, ${\it mgt}_t \in \{0, 1\}$ is the binary state of the magnet, $\mathbf{ef}_{\mathit{xyz}, t} \in \mathbb{R}^3$ and $\mathbf{ef}_{R, t} \in \mathbb{H}$ are the position and the orientation of the robot's end-effector, $\mathbf{g}_{\mathit{xyz}, t} \in \mathbb{R}^3$ and $\mathbf{g}_{R, t} \in \mathbb{H}$ are the position and the orientation of the goal object, and $\mathbf{o}_{\mathit{xyz}, t} \in \mathbb{R}^3$ and $\mathbf{o}_{R, t} \in \mathbb{H}$ are the position and the orientation of the obstacle object at time $t$.

The action vector $\mathbf{a}_t$ leading from $\mathbf{s}_t$ to $\mathbf{s}_{t + 1}$ is given as
\begin{equation}
    \mathbf{a}_t \triangleq \left(\Delta \boldsymbol{\theta}_t, \Delta {\it mgt}_t \right),
    \label{eq:action_vector}
\end{equation}
where $\Delta \boldsymbol{\theta}_t \triangleq \boldsymbol{\theta}_{t + 1} - \boldsymbol{\theta}_t$, and $\Delta {\it mgt}_t \triangleq {\it mgt}_{t + 1} - {\it mgt}_t$.

The dataset generation process was accomplished in a simulated environment using a MyGym toolkit \cite{Vavrecka2021}. The robotic manipulator was a KUKA LBR iiwa with 7 degrees of freedom and equipped with a magnet attached to the last link instead of a gripper. The environment contained a single static obstacle object, a primitive oriented box. A special pose of the obstacle, $\mathbf{o}_{\emptyset} \triangleq [{(-2)}_{\times 7}]$, was selected arbitrarily to represent no obstacle present in the environment, as such a position lies outside the valid workspace and thus does not correspond to any physically realisable obstacle pose.

The simulation was used to generate three training datasets. Datasets $\mathcal{D}_1$ and $\mathcal{D}_2$ consisted of 500\,000 transitions, both with the ratio of transitions in the obstacle vs. obstacle-free setups at 9:1, while the ratio of colliding to non-colliding transitions was 4:1. These ratios were chosen to increase the density of samples in regions where the dynamics are more complex and critical for learning. $\mathcal{D}_1$ contained multiple transitions per setup, contrasting $\mathcal{D}_2$ with only a single transition per setup. The simulation was also used to generate training trajectories. Similarly, the generation was done in both obstacle-containing and obstacle-free environments. A ratio similar to that for transitions was enforced. This produced a set $\mathcal{D}_3$ of 12\,000 trajectories with a maximum allowed length of 52 states.

The FM and IM were obtained by the same training process as described in \cite[Sec.~5.2]{Krupa2026}. The former was trained on the dataset $\mathcal{D}_1$, while the latter was trained using $\mathcal{D}_2$. The evaluations of the FM and IM are presented in Tab.~\ref{tab:fm_im_test}. The models were evaluated on multiple metrics, all measured as mean absolute errors except the rotational subvectors, which were computed as mean geodesic error (i.e., the angular distance between two rotations on the unit quaternion sphere) to better reflect the quaternion representation.

\begin{table}[t]
    \centering
    \scriptsize
    \caption[FM and IM test performance across transition categories]{FM and IM test performance across transition categories.  Mean absolute error was measured for the corresponding subvector of the state/action vector. Test set contained 30\,000 transitions, evenly divided into 3 categories: no obstacle present, non-colliding in the presence of an obstacle, and resulting in a collision.}
    \label{tab:fm_im_test}
    \begin{tabular}{@{}lccccccccc@{}}
        \toprule
         & \multicolumn{7}{c}{Forward model} & \multicolumn{2}{c}{Inverse model} \\ \cmidrule(lr){2-8}\cmidrule(lr){9-10} 
         &
          $\boldsymbol{\theta}$ [rad] &
          $\mathbf{ef}_{\it xyz}$ [m] &
          $\mathbf{ef}_R$ [rad] &
          $\mathbf{g}_{\it xyz}$ [m] &
          $\mathbf{g}_R$ [rad] &
          $\mathbf{o}_{\it xyz}$ [m] &
          $\mathbf{o}_R$ [rad] &
          $\Delta \boldsymbol{\theta}$ [rad] &
          $\Delta {\it mgt}$ \\ \midrule
        No obstacle   & 0.015 & 0.010 & 0.070 & 4.4e-4 & 9.8e-4 & 1.9e-4 & 1.0e-3 & 0.0143          & 2.6e-5          \\
        Colliding     & 0.057 & 0.020 & 0.134 & 1.6e-4 & 9.8e-4 & 1.6e-4 & 1.1e-3 & 0.0313          & 9.0e-6          \\
        Non-colliding & 0.016 & 0.010 & 0.072 & 3.5e-4 & 9.8e-4 & 3.3e-4 & 1.4e-3 & 0.0139          & 2.4e-5          \\ \bottomrule
    \end{tabular}
\end{table}

With the supervisory models trained, the TM architecture and training were examined. The loss function $\mathcal{L}_{\rm TM}$ from Eq.~\ref{eq:LTM} was minimised with all terms using the same loss functions as for the FM architecture: $\mathcal{L}_{\boldsymbol{\theta}}$ (joint configuration error) and $\mathcal{L}_{\it xyz}$ (total error of positional outputs $\mathbf{ef}_{\it xyz}$, $\mathbf{g}_{\it xyz}$, $\mathbf{o}_{\it xyz}$) use mean squared error of the corresponding subvectors, while $\mathcal{L}_{\it mgt}$ uses binary cross-entropy with logits. Lastly, we define the loss function $\mathcal{L}_{R}$ for the rotational heads as
\begin{equation}
    \mathcal{L}_{R} = \mathcal{L}_{\mathrm{chordal}} \!\left(\mathbf{\hat{y}}_R, \mathbf{y}_R\right) = 1 - \left| \mathbf{\hat{y}}_R^{\top} \mathbf{y}_R \right|.
    \label{eq:Lchordal}
\end{equation} 

The initial and final states from the trajectory training dataset served as input during the experiments, and their corresponding loss functions were computed as
\begin{align}
    \mathcal{L}_{\mathrm{init}} \!\left(\mathbf{\hat{s}}_1, \mathbf{s}_0 \right) &=
        \mathcal{L}_{\boldsymbol{\theta}} \!\left(\boldsymbol{\hat{\theta}}_1, \boldsymbol{\theta}_0 \right) + \mathcal{L}_{\it xyz} \!\left(\mathbf{\hat{ef}}_1, \mathbf{ef}_0 \right) \\
    \mathcal{L}_{\mathrm{goal}} \!\left(\mathbf{\hat{s}}_{T - 1}, \mathbf{s}_T \right) &=
        \mathcal{L}_{\boldsymbol{\theta}} \!\left(\boldsymbol{\hat{\theta}}_{T - 1}, \boldsymbol{\theta}_T \right) + \mathcal{L}_{\it xyz} \!\left(\mathbf{\hat{ef}}_{T - 1}, \mathbf{ef}_T \right)
\end{align}

\begin{figure}[ht]
    \centering
    \includegraphics[width=\textwidth]{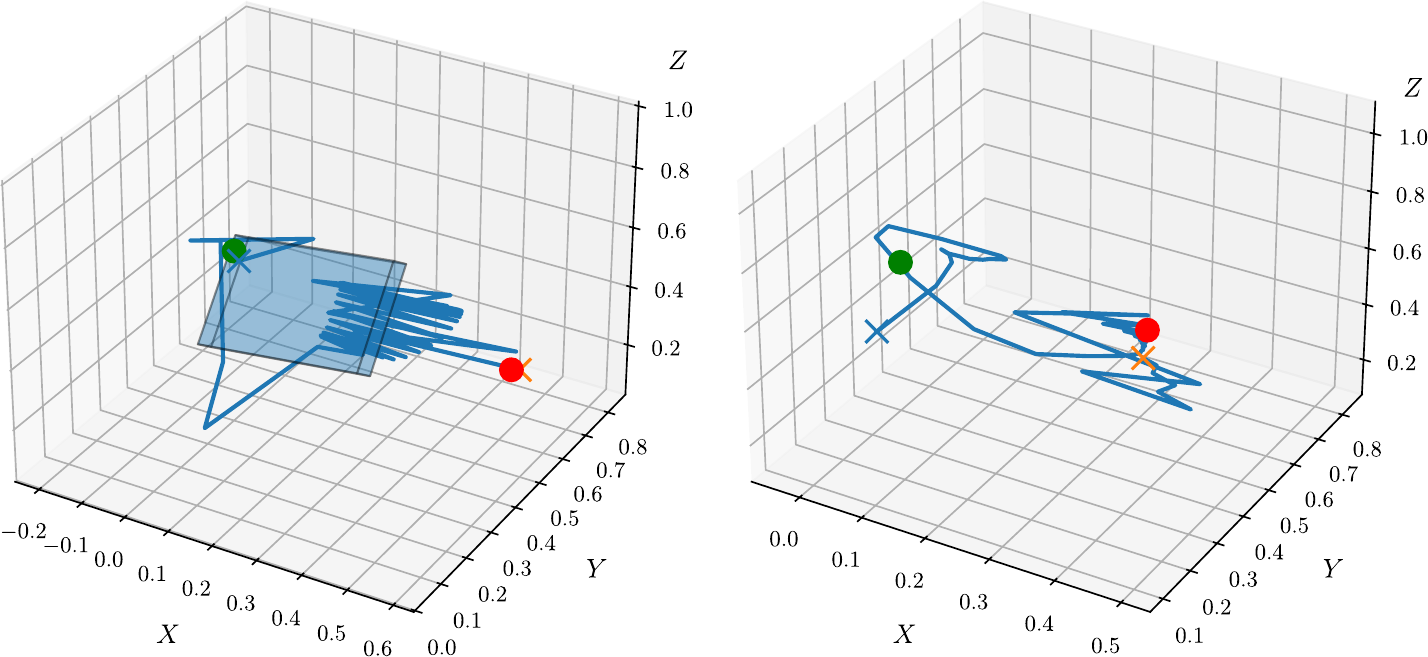}
    \caption[3D visualisations of generated trajectories by TM exploiting rectification]{Generated trajectories' waypoints in a space with (left) and without an obstacle for a model exploiting the rectification learning feedback. \emph{Green} and \emph{red points} mark the initial and the final ground-truth state, respectively. The planned trajectory starts in the \emph{blue crossmark} and terminates in the \emph{orange} one. Oscillation pattern in safe regions is learned and repeated for every trajectory.}
    \label{fig:tm_exploit}
\end{figure}

The initial attempts at hyperparameter tuning with weaker internal models exposed a weakness of this framework \cite[Sec.~6.2]{Krupa2026}, where a model with large enough capacity learned to exploit the rectification learning signal, creating trajectories, while consistent with the FM and IM, that performed unnatural and unnecessary movement in safe areas, where action and its resulting state were easily predictable. The model produced paths oscillating in these regions as long as possible to lower the rectification loss, after which heaping movements towards the goal followed to minimise the penalty for distance away from the goal position (see Fig.~\ref{fig:tm_exploit}). 

Various prevention mechanisms were tested to mitigate the exploitation, including supervised pretraining based on ground-truth trajectories from the training dataset and better tuning of the rectifying models. The next approach introduced additional end-effector-based geometric priors through a separate geometric loss term encouraging smoother spacing between waypoints and more geometrically consistent trajectories by preventing large steps, sharp angles, unnecessarily long trajectories, and oscillations. 

These modifications significantly improved trajectory quality and reduced oscillatory behaviour during training. We constructed 5 different models that were further used for experimentation. Each model was trained using the AdamW optimiser \cite{Loshchilov2019} with the initial learning rate $\eta= 1 \times 10^{-4}$. To prevent gradient explosion for long-horizon recurrent model predictors, a horizon of $n_{\rm timesteps} = 50$ was chosen, and gradient clipping was implemented with a clip norm of $2.0$ to stabilise the learning curves. 

\begin{figure}[t]
    \centering
    \includegraphics[width=\textwidth]{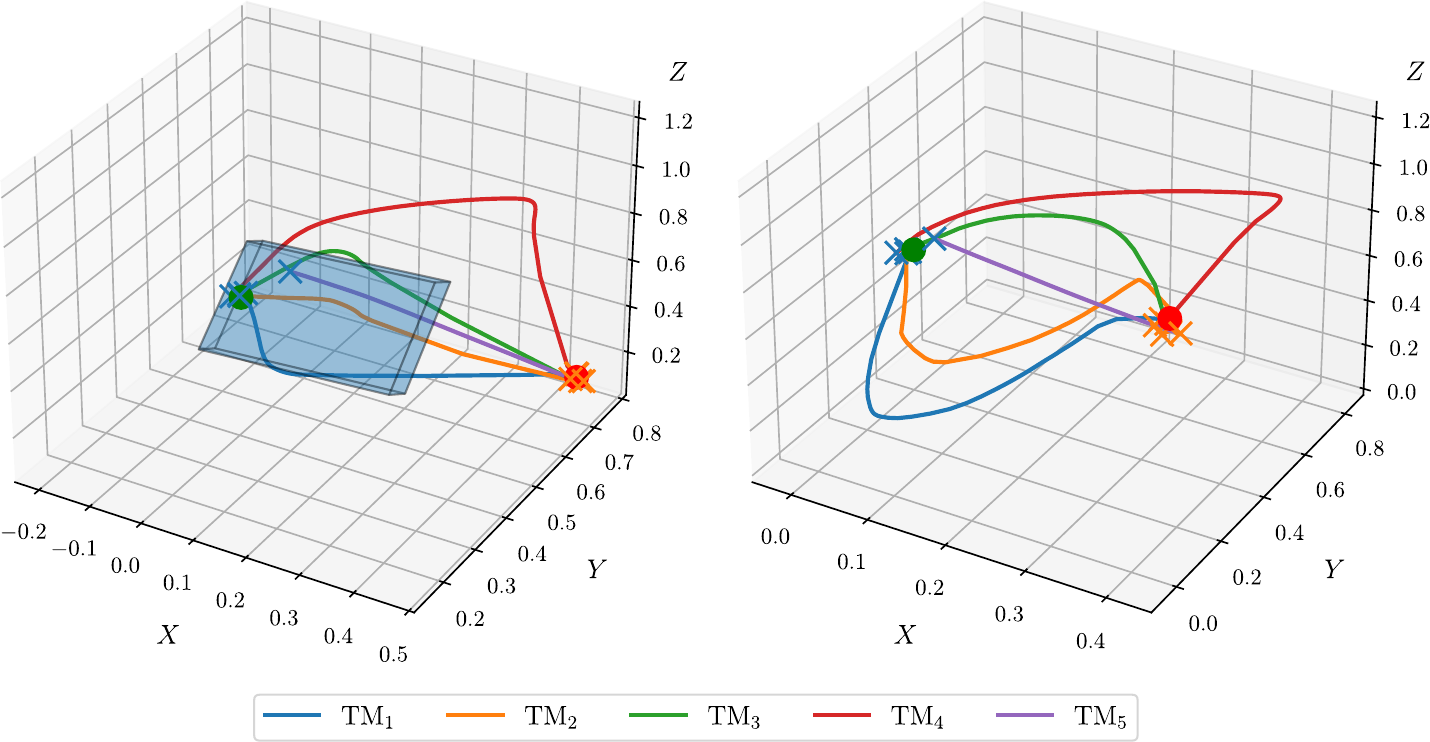}
    \caption[3D visualisations of generated trajectories by TM with negated exploitation]{Generated trajectories' waypoints in a space with (left) and without an obstacle for models after exploitation prevention. \emph{Green} and \emph{red points} mark the initial and the final ground-truth state, respectively. The planned trajectories start in the \emph{blue crossmarks} and terminate in the \emph{orange} ones.}
    \label{fig:tm_trajs}
\end{figure}

The trajectory models were evaluated in two phases. The first phase focused on the geometric properties of the generated trajectories, assessed based on the distance to the initial position, distance to the goal position, transition sizes, average angle between consecutive trajectory segments, and tail-related metrics, i.e., the real effective length of the predicted trajectory and what fraction of available steps goes unused. The distances to the boundary positions indicate whether the models can correctly position trajectories in the workspace, with the goal-position distance specifically showing a model's ability to correctly perform trajectory planning. Both step sizes and angles between the points determine the smoothness and geometric consistency of the trajectories. Large variations between individual step sizes or sharp turns lead to unnatural and inefficient motions. All metrics are calculated based on generated trajectories' end-effector positions $\hat{\tau}_{\it ef}$. An additional test set of 4\,000 trajectories was generated, with 2\,000 containing no obstacle in the environment. Their initial and goal states were used for querying the model. During training, each architecture was evaluated every 10 epochs. 

Although the models learn how to generate trajectories with higher geometric consistency, successful execution in the simulated environment additionally depends on correct timing and physically feasible positioning. We evaluated the models' ability to generate practically executable trajectories by executing the predictions directly in the simulator. As an additional metric, we measured the ground-truth distance to the goal position and the differences between predicted and executed trajectories to assess whether the generated waypoints were correctly followed. Furthermore, the ratio of successful trajectories and collisions was recorded. Execution was evaluated in two modes: following the generated configurations directly and following configurations obtained via inverse kinematics using the generated end-effector positions. A summary of this experiment is presented in Table~\ref{tab:tm_sim}.

\begin{table}[t]
\centering
\scriptsize
\setlength{\tabcolsep}{3.5pt}

\caption[Simulator evaluation metrics of the trajectory models]{Simulator evaluation metrics of the trajectory models measured on the generated test trajectory set. The table reports collision rate $c\%$, raw magnet success rate $m_{\it raw}\%$, collision-free magnet success rate $m\%$, waypoint following rate $w\%$, average number of repeated control steps per waypoint $r$, final distance to goal $d_{\it goal}$, mean executed step size $\bar{s}$, mean executed trajectory angle $\bar{\alpha}$, and average deviation between predicted and executed trajectories $d_{\it exec}$. Lower values are better for $c\%$, $r$, $d_{\it goal}$, $\bar{s}$, and $d_{\it exec}$, while higher values are better for $m_{\it raw}\%$, $m\%$, and $w\%$. Each model, mode, and environmental setup was tested with 200 episodes. Entries marked with ``--'' indicate that the metric could not be computed due to an inverse kinematics failure.}
\label{tab:tm_sim}

\begin{tabular}{@{}llcccc@{}}
\toprule
& & $c\%$ [\%] & $m_{\it raw}\%$ [\%] & $m\%$ [\%] & $w\%$ [\%] \\
\cmidrule(l){3-6}
Model & Mode & No obs. / Obs. & No obs. / Obs. & No obs. / Obs. & No obs. / Obs. \\
\midrule
\multirow{2}{*}{$\mathrm{TM}_1$}
& Cfg. &
0.0 / 14.5 &
100.0 / 86.0 &
100.0 / 85.5 &
99.0 / 98.2 \\
& EE+IK &
-- / 43.0 &
-- / 78.0 &
-- / 57.0 &
-- / 0.0 \\
\midrule
\multirow{2}{*}{$\mathrm{TM}_2$}
& Cfg. &
0.0 / 44.5 &
100.0 / 84.5 &
100.0 / 55.5 &
88.8 / 79.4 \\
& EE+IK &
0.0 / 44.0 &
58.0 / 80.0 &
58.0 / 56.0 &
0.0 / 0.0 \\
\midrule
\multirow{2}{*}{$\mathrm{TM}_3$}
& Cfg. &
0.0 / 16.5 &
100.0 / 84.5 &
100.0 / 82.0 &
63.7 / 64.9 \\
& EE+IK &
0.0 / 46.5 &
100.0 / 81.0 &
100.0 / 53.0 &
0.0 / 0.0 \\
\midrule
\multirow{2}{*}{$\mathrm{TM}_4$}
& Cfg. &
0.0 / 20.0 &
100.0 / 89.5 &
100.0 / 80.0 &
63.0 / 61.7 \\

& EE+IK &
0.0 / 61.5 &
99.5 / 61.0 &
99.5 / 37.0 &
0.1 / 0.0 \\
\midrule
\multirow{2}{*}{$\mathrm{TM}_5$}
& Cfg. &
0.0 / 7.5 &
100.0 / 95.5 &
100.0 / 92.5 &
99.8 / 96.7 \\
& EE+IK &
0.0 / 47.0 &
97.0 / 79.0 &
97.0 / 51.0 &
0.0 / 0.0 \\
\bottomrule
\end{tabular}

\vspace{\belowcaptionskip}

\begin{tabular}{@{}llccccc@{}}
\toprule
& & $r$ & $d_{\it goal}$ [m] & $\bar{s}$ [m] & $\bar{\alpha}$ [$^\circ$] & $d_{\it exec}$ [m] \\
\cmidrule(l){3-7}
Model & Mode & No obs. / Obs. & No obs. / Obs. & No obs. / Obs. & No obs. / Obs. & No obs. / Obs. \\
\midrule
\multirow{2}{*}{$\mathrm{TM}_1$}
& Cfg. &
2.02 / 2.09 &
0.131 / 0.177 &
0.096 / 0.082 &
148.92 / 159.75 &
0.725 / 0.677 \\
& EE+IK &
-- / 20.0 &
-- / 0.185 &
-- / 0.025 &
-- / 172.78 &
-- / 0.151 \\
\midrule
\multirow{2}{*}{$\mathrm{TM}_2$}
& Cfg. &
11.27 / 12.54 &
0.134 / 0.178 &
0.508 / 0.430 &
30.47 / 18.91 &
0.719 / 0.807 \\
& EE+IK &
11.6 / 20.0 &
0.175 / 0.185 &
0.037 / 0.029 &
168.71 / 172.49 &
0.131 / 0.158 \\
\midrule
\multirow{2}{*}{$\mathrm{TM}_3$}
& Cfg. &
11.39 / 10.95 &
0.146 / 0.189 &
0.389 / 0.391 &
12.87 / 12.29 &
0.796 / 0.808 \\
& EE+IK &
20.0 / 19.9 &
0.132 / 0.180 &
0.029 / 0.029 &
173.89 / 171.95 &
0.129 / 0.181 \\
\midrule
\multirow{2}{*}{$\mathrm{TM}_4$}
& Cfg. &
11.87 / 11.70 &
0.128 / 0.163 &
0.433 / 0.427 &
12.25 / 11.28 &
0.827 / 0.869 \\
& EE+IK &
19.9 / 19.7 &
0.148 / 0.225 &
0.039 / 0.033 &
171.98 / 167.44 &
0.130 / 0.297 \\
\midrule
\multirow{2}{*}{$\mathrm{TM}_5$}
& Cfg. &
1.05 / 1.63 &
0.128 / 0.140 &
0.016 / 0.017 &
166.61 / 165.20 &
0.144 / 0.149 \\
& EE+IK &
19.4 / 19.6 &
0.150 / 0.183 &
0.015 / 0.016 &
175.41 / 167.63 &
0.131 / 0.171 \\
\bottomrule
\end{tabular}
\end{table}

\section{Discussion and Future Work}
\label{discussion}

Our results demonstrate that self-supervised recurrent trajectory planning can generate feasible and geometrically meaningful robotic motion in static environments. At the same time, the experiments revealed limitations of the framework and highlighted directions for future improvement.

We observed the TM's tendency to exploit the rectification mechanism. TMs with larger capacity gradually learned to minimise the rectification loss without generating meaningful trajectories. Instead of naturally spacing the intermediate states between the initial and goal positions, they often generated oscillatory or near-stationary trajectories that locally minimised the rectification error produced by the FM and IM. This points towards the TM effectively overfitting to imperfections of the self-supervised feedback mechanism. Since the rectification process depends completely on the learned FM and IM, inaccuracies of these models create regions in the state-action space where unrealistic transitions can still achieve low rectification loss. As the TM capacity increased, the model became capable of finding and exploiting such regions. 

Both geometric priors and supervised pretraining had similar effects on performance. In both cases, the model first learned geometrically meaningful trajectories before the rectification process became the main learning objective. This suggests that such training strategies may provide a more robust mitigation mechanism against exploitation of the self-supervised feedback process. However, the experiments suggest that exploiting learned supervisory models remains one of the main limitations of the proposed self-supervised framework, highlighting the importance of the FM and IM optimisation.

We hypothesise that geometric consistency is not sufficient for the successful execution of trajectories. Several models achieved relatively good geometric metrics and low rectification errors, but still planned trajectories that were difficult to execute. This was especially visible during the initial experiments without action repetition, where the self-supervised models struggled to correctly follow the generated waypoints. This suggests that even small local prediction errors can accumulate during execution and cause divergence from the originally planned trajectory. The difference between generated and executed trajectories also indicates that the TMs learned an approximation of the dynamics represented by the internal models instead of the true simulator dynamics. Since the FM and IM themselves contain prediction errors, the TM optimises for consistency with these approximations. As a result, a low rectification error does not necessarily guarantee physically feasible execution inside the simulator.
The comparison between $\mathrm{TM}_2$, $\mathrm{TM}_3$, and $\mathrm{TM}_4$ showed that both geometric priors and supervised pretraining significantly improved the behaviour of the larger trajectory models in the no-obstacle environment. $\mathrm{TM}_2$, trained solely through the self-supervised rectification, frequently generated trajectories that did not lead to the successful completion of the task. This suggests that a more guided optimisation approach helps the model learn in an obstacle-free setup.%

Most promising is the smaller $\mathrm{TM}_1$ architecture. Despite its smaller capacity, $\mathrm{TM}_1$ consistently achieved better execution quality in obstacle environments in both execution modes. Compared to the larger self-supervised networks, $\mathrm{TM}_1$ achieved significantly better collision rate, success rate, waypoint reach rate, and required fewer repeated actions during execution. The model also generated smoother trajectories and fewer unrealistic transitions. In several obstacle-related metrics, $\mathrm{TM}_1$ achieved better performance than the fully supervised $\mathrm{TM}_5$ model. This behaviour suggests that with a smaller representation space, the model was forced to economise, choosing the simplest solution to the problem, which led to implicit learning of what a good-quality trajectory respecting the environmental constraints looks like. As a result, the smaller recurrent architecture generalised better and was less capable of exploiting the weakness of the self-supervised rectification mechanism. Therefore, smaller models may be more favourable for obstacle-aware planning, as they have lower spatial and temporal complexity in both training and inference. However, even the $\mathrm{TM}_1$ architecture did not produce executable trajectories in all environments and conditions.

Ultimately, all self-supervised trajectory models' executed trajectories diverged from the originally planned motion. This further emphasises the importance of accurate FM and IM, as the TM optimises consistency with approximated dynamics instead of true simulator dynamics. We will direct our future work toward improving the quality of the internal models to improve executability and long-horizon trajectory consistency. The obtained results indicate that self-supervised recurrent trajectory planning represents a promising direction for robotic motion planning. The current limitations, hypothesised to be related to learned dynamics approximation quality and exploitation, will be addressed through improved internal models, richer datasets, stronger geometric constraints, and more advanced architectures.

\subsubsection{Acknowledgements.}
The authors thank Matthias Kerzel (Department of Informatics, University of Hamburg) for his advice and rigorous feedback. This research was supported by the Slovak Research and Development Agency, project APVV-21-0105. Research results were in part obtained using the computational resources of the supercomputer PERUN at the Supercomputing Center at TU Košice, with support from the EU, funds of the Recovery and Resilience Plan of the Slovak Republic, project 17I03-04-P03-00001. We also thank the Slovak Society for Cognitive Science (SSKV)\footnote{\url{https://cogsci.fmph.uniba.sk/sskv/}} for their support.

\bibliographystyle{splncs04}
\bibliography{references} 

@Article{Karaman2011,
  author    = {Karaman, Sertac and Frazzoli, Emilio},
  journal   = {The Int. Jour. of Robotics Research},
  title     = {Sampling-based algorithms for optimal motion planning},
  year      = {2011},
  issn      = {1741-3176},
  number    = {7},
  pages     = {846--894},
  volume    = {30},
  doi       = {10.1177/0278364911406761},
  publisher = {SAGE Publications},
}

@Article{Xie2019,
  author    = {Xie, Jiexin and Shao, Zhenzhou and Li, Yue and Guan, Yong and Tan, Jindong},
  journal   = {IEEE Access},
  title     = {Deep reinforcement learning with optimized reward functions for robotic trajectory planning},
  year      = {2019},
  issn      = {2169-3536},
  pages     = {105669--105679},
  volume    = {7},
  doi       = {10.1109/access.2019.2932257},
  publisher = {Institute of Electrical and Electronics Engineers (IEEE)},
}

@MastersThesis{Krupa2026,
  author = {Krupa, Miroslav},
  school = {Comenius University Bratislava},
  title  = {Self-Supervised Robotic Trajectory Planning with Obstacle Avoidance},
  year   = {2026},
  type   = {Bachelor's thesis},
}

@Article{FigueiredoPrudencio2024,
  author    = {Figueiredo Prudencio, Rafael and Maximo, Marcos R. and Colombini, Esther Luna},
  journal   = {IEEE Trans. on Neural Networks and Learning Systems},
  title     = {A survey on offline reinforcement learning: {T}axonomy, review, and open problems},
  year      = {2024},
  issn      = {2162-2388},
  number    = {8},
  pages     = {10237--10257},
  volume    = {35},
  doi       = {10.1109/tnnls.2023.3250269},
  publisher = {Institute of Electrical and Electronics Engineers (IEEE)},
}

@InProceedings{Emmons2022,
  author    = {Scott Emmons and Benjamin Eysenbach and Ilya Kostrikov and Sergey Levine},
  booktitle = {Int. Conf. on Learning Representations},
  title     = {{RvS}: {W}hat is essential for offline {RL} via supervised learning?},
  year      = {2022},
  url       = {https://openreview.net/forum?id=S874XAIpkR-},
}

@Article{Hu2026,
  author    = {Hu, Jia and Chang, Yang and Wang, Haoran},
  journal   = {Transportation Research Part C: Emerging Technologies},
  title     = {A review of learning-based motion planning: {T}oward a data-driven optimal control approach},
  year      = {2026},
  issn      = {0968-090X},
  pages     = {105767},
  volume    = {190},
  doi       = {10.1016/j.trc.2026.105767},
  publisher = {Elsevier BV},
}

@Article{ElgueaAguinaco2024,
  author    = {Elguea-Aguinaco, {{\'{I}}}{\~{n}}igo and Inziarte-Hidalgo, Ibai and B{\o}gh, Simon and Arana-Arexolaleiba, Nestor},
  journal   = {Int. Journal of Intel. Systems},
  title     = {A review on reinforcement learning for motion planning of robotic manipulators},
  year      = {2024},
  issn      = {1098-111X},
  number    = {1},
  volume    = {2024},
  date      = {2024},
  doi       = {10.1155/int/1636497},
  editor    = {Khosravi, Mohamadreza (Mohammad)},
  publisher = {Wiley},
}

@Article{Dobis2022,
  author    = {Dobiš, Michal and Dekan, Martin and Beňo, Peter and Duchoň, František and Babinec, Andrej},
  journal   = {Robotics},
  title     = {Evaluation criteria for trajectories of robotic arms},
  year      = {2022},
  issn      = {2218-6581},
  number    = {1},
  volume    = {11},
  doi       = {10.3390/robotics11010029},
  publisher = {MDPI AG},
}

@InProceedings{Janner2021,
  author    = {Janner, Michael and Li, Qiyang and Levine, Sergey},
  booktitle = {Advances in NeurIPS},
  title     = {Offline reinforcement learning as one big sequence modeling problem},
  year      = {2021},
  pages     = {1273--1286},
  volume    = {34},
}

@InProceedings{Cibula2025,
  author    = {Cibula, Miroslav and Malinovská, Kristína and Kerzel, Matthias},
  booktitle = {Int. Conf. on Artif. Neural Networks},
  title     = {Towards bio-inspired robotic trajectory planning via self-supervised {RNN}},
  year      = {2025},
  pages     = {149--160},
  publisher = {Springer Nature},
  doi       = {10.1007/978-3-032-04552-2_15},
  isbn      = {978-3-032-04552-2},
  issn      = {1611-3349},
}

@InProceedings{Vavrecka2021,
  author    = {Vavrečka, Michal and Sokovnin, Nikita and Mejdrechová, Megi and Šejnová, Gabriela},
  booktitle = {33rd Int. Conf. on Tools with AI ({ICTAI})},
  title     = {{MyGym}: {Modular} toolkit for visuomotor robotic tasks},
  year      = {2021},
  pages     = {279--283},
  publisher = {IEEE},
  doi       = {10.1109/ictai52525.2021.00046},
}

@InProceedings{Cho2014,
  author    = {Cho, Kyunghyun and van Merrienboer, Bart and Gulcehre, Caglar and Bahdanau, Dzmitry and Bougares, Fethi and Schwenk, Holger and Bengio, Yoshua},
  booktitle = {2014 Conf. on Empirical Methods in NLP (EMNLP)},
  title     = {Learning phrase representations using {RNN} encoder--decoder for statistical machine translation},
  year      = {2014},
  pages     = {1724--1734},
  doi       = {10.3115/v1/d14-1179},
}

@InProceedings{Loshchilov2019,
  author    = {Ilya Loshchilov and Frank Hutter},
  booktitle = {Int. Conf. on Learning Representations},
  title     = {Decoupled weight decay regularization},
  year      = {2019},
  url       = {https://openreview.net/forum?id=Bkg6RiCqY7},
}

\end{document}